\documentclass{article}
\usepackage[T1]{fontenc}

\usepackage[preprint]{arxiv}
\usepackage{amsmath,amssymb}
\usepackage{amsthm}
\usepackage{graphicx}
\usepackage{subcaption}
\usepackage{booktabs}
\usepackage{multirow}
\usepackage{tabularx}
\usepackage{threeparttable}
\usepackage{tikz}
\usetikzlibrary{arrows.meta,positioning}
\usepackage{pifont}
\usepackage{hyperref}
\usepackage[capitalize,noabbrev]{cleveref}
\usepackage{enumitem}

\title{GestaltBench: Can Vision-Language Models Integrate Local\\ Cues into Global Percepts?}
\headl{GestaltBench}
\headr{Yuanjun Pu et al.}

\emails{tianlin001@buaa.edu.cn}

\author{%
  \mbox{Yuanjun Pu\textsuperscript{1}\cofirst}
  \And
  \mbox{Moran Sun\textsuperscript{2}\cofirst}
  \And
  \mbox{Zhiheng Zhang\textsuperscript{3}}
  \And
  \mbox{Xiaoyu Zhang\textsuperscript{4}}
  \And
  \mbox{Tianlin Li\textsuperscript{1}\correspond}
  \And
  \mbox{Yihao Huang\textsuperscript{5}}
  \And
  \mbox{Guangyue Sun\textsuperscript{2}}
  \And
  \mbox{Yuanshuai Wang\textsuperscript{1}}
  \And
  \mbox{Jinkun Zhao\textsuperscript{1}}
  \And
  \mbox{Kunwei Wang\textsuperscript{3}}
  \And
  \mbox{Ruibo Chen\textsuperscript{1}}
}
\affiliations{%
  \mbox{\textsuperscript{1}Beihang University}
  \quad
  \mbox{\textsuperscript{2}Nanyang Technological University}
  \quad
  \mbox{\textsuperscript{3}RWTH Aachen University} \\
  \mbox{\textsuperscript{4}The Hong Kong University of Science and Technology}
  \quad
  \mbox{\textsuperscript{5}East China Normal University}
}

\begin{document}

\maketitle

\begin{abstract}


Despite strong visual recognition abilities, modern vision-language models (VLMs) struggle with \textbf{holistic perceptual organization}, the ability to integrate local visual elements into coherent global percepts. 
Drawing on Gestalt psychology, which provides the theoretical basis for such perceptual phenomena, we introduce \textbf{GestaltBench}, a benchmark comprising 3,613 instances across 12 PO task families and 1 reference set based on four key Gestalt principles: reification, emergence, invariance, and multistability. 
Extensive evaluations of 11 VLMs reveal substantial limitations in this capability: models average only 18.68\% macro-averaged accuracy across the PO task families (best: 29.69\%), compared with 78.39\% on the reference family and 66.43\% for human participants.
Building on these findings, we develop a proof-of-concept Gestalt-based CAPTCHA for Web human verification. On 16 held-out challenges, humans achieve 66.07\% exact-set accuracy, while three VLMs average just 20.83\%; this substantial Human--VLM gap is a prerequisite for human verification, indicating the Web-security potential of GestaltBench.
Code and data are available at \url{https://github.com/LambdaZ/GestaltBench}.

\end{abstract}

\section{Introduction}

Vision-language models (VLMs) excel at image captioning, scene-text recognition, and visual reasoning~\cite{lin2014microsoft,mathew2021docvqa,li2023seed}. However, they often struggle with \emph{holistic perceptual organization} (PO), the ability to integrate fragmented visual cues into coherent global percepts.
For example, humans can infer words from dot patterns, recognize animals from sparse contours, and identify people in abstract portraits, even when the visual cues are incomplete or unconventional. 
Interpreting such images involves more than looking at separate local visual parts. It requires combining fragmentary cues into a coherent global view to uncover what the image is meant to convey.

\begin{figure}[t]
\centering
\begin{tikzpicture}[font=\footnotesize,>=Stealth]
  \node (img) {\includegraphics[width=2.0cm]{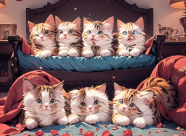}};
  \node[align=center,below=3.5mm of img,xshift=-2.05cm] (vlm)
    {\textbf{VLM}\\[1pt]\textcolor{red!75!black}{  \emph{carrier} \ding{55}}};
  \node[draw=red!55,fill=red!5,rounded corners,inner sep=2pt,below=1.2mm of vlm,text width=2.8cm,align=center,font=\scriptsize] (vlmans)
    {``a group of cats on couches\,''};
  \node[align=center,below=3.5mm of img,xshift=2.05cm] (hum)
    {\textbf{Human}\\[1pt]\textcolor{green!50!black}{recognizes \emph{target} \ding{51}}};
  \node[draw=green!55!black,fill=green!7,rounded corners,inner sep=2pt,below=1.2mm of hum,text width=2.8cm,align=center,font=\scriptsize] (humans)
    {``\texttt{SEND NUDES}''};
  \draw[->] (img) -- (vlm.north);
  \draw[->] (img) -- (hum.north);
\end{tikzpicture}

\caption{Semantic target recognition. The visible carrier depicts cats, while the intended target is \texttt{SEND NUDES}. A carrier-only description is not counted as recognition.}
\label{fig:overview}
\end{figure}

\begin{figure*}[t]
  \centering
  \includegraphics[width=0.84\textwidth]{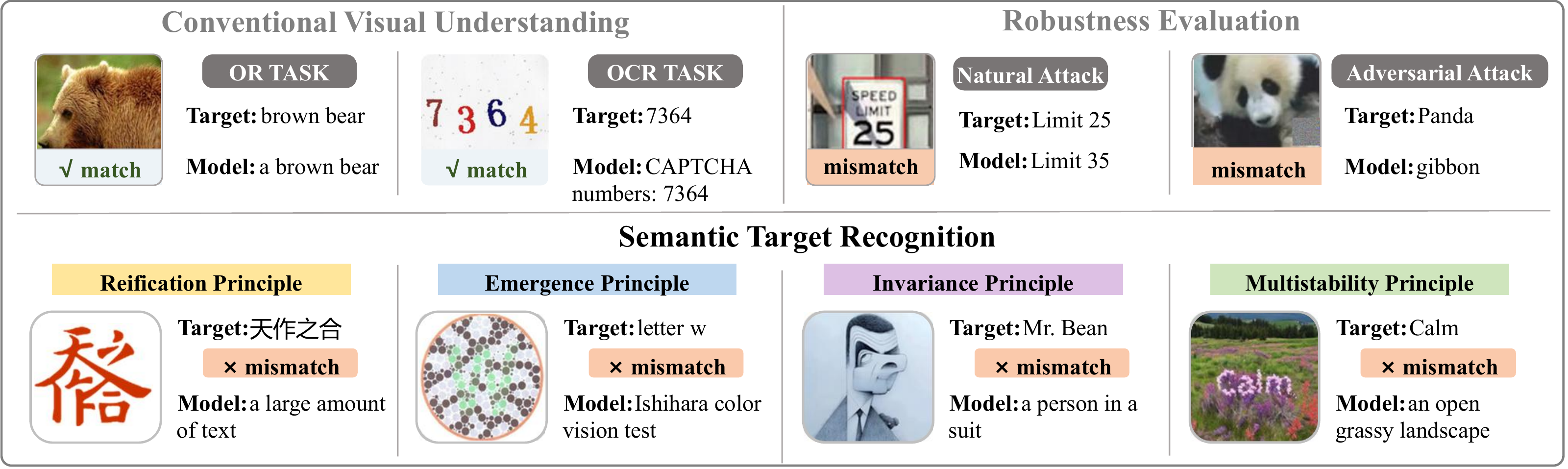}
  \caption{Visual understanding and robustness retain explicit targets; GestaltBench evaluates recognition of a human-recognizable target deliberately concealed by its carrier.}
  \label{fig:teaser}
\end{figure*}

Systematically evaluating holistic perceptual organization remains challenging because existing benchmarks largely focus on extracting an image’s dominant interpretation, rather than recognizing an intended target that differs from its most salient visual reading. 
For example, some datasets evaluate text recognition and visual reasoning~\cite{liu2024mmbench, liu2024ocrbench} or test whether predictions remain correct under image corruptions or misleading questions~\cite{hendrycks2019benchmarking,liu2024seeing}; further work examines whether model descriptions hallucinate~\cite{li2023evaluating}.
These evaluations primarily assess general visual understanding and answer reliability, but provide limited systematic assessment of semantic target recognition from incomplete, transformed, or embedded visual cues.
Existing studies have examined perceptual organization in models, with synthetic grouping tasks and visual illusions~\cite{qi2021pqa,zhang2025illusionbench,shahgir2024illusionvqachallengingopticalillusion}; but they cover limited selected task families, leaving broader recognition demands underexplored.

To address this, we introduce GestaltBench, a diagnostic benchmark comprising 3,613 instances across 12 PO task families and an ordinary-animal (OA) reference set.
Drawing on Gestalt psychology, the PO families are grouped by their dominant recognition operations into four categories: \emph{reification}, \emph{emergence}, \emph{invariance}, and \emph{multistability}.
Task design emphasizes global organization of visual cues and human recognizability of the target.
Images are constructed or curated through four routes: inherited transformation, scripted construction, template-conditioned diffusion, and collection and curation.
Manual quality screening removes unrecognizable or visually defective images, while annotation fixes each target and its accepted aliases set before evaluation.
Target-recognition accuracy serves as the primary score, with auxiliary scores for specific families probing residual information when complete recognition fails.

Extensive evaluations of 11 VLMs reveal substantial limitations in semantic target recognition: models average only 18.68\% macro-averaged accuracy across the 12 PO task families (best: 29.69\%), compared with 78.39\% mean accuracy on the OA reference set, while seven human participants achieve 66.43\% on sampled items.
Building on these findings, we develop a proof-of-concept Gestalt-based selection CAPTCHA. On 16 held-out challenges, humans achieve 66.07\% mean exact-set accuracy, while three VLMs average just 20.83\%.
As advanced VLMs have posed challenges to image-based human verification on the Web~\cite{309468}, this 45.24-pp Human--VLM gap highlights the Web-security potential of PO tasks in GestaltBench.

\noindent\textbf{Contributions.} Our contributions are threefold: (1) we introduce GestaltBench, a benchmark for systematically assessing semantic target recognition; (2) we conduct extensive evaluations across 11 VLMs, revealing substantial performance gaps, mechanism-specific variations, and persistent blind spots; and (3) we demonstrate the practical Web-security utility of our findings through a proof-of-concept Gestalt-based CAPTCHA for human verification.



\section{Related Work}
\label{sec:prelim-related}


\textbf{Visual Understanding and Reliability.} 
General VLM benchmarks examine models from three complementary perspectives.
\textit{Capability benchmarks} measure whether models can read and reason over visual content, ranging from captioning, scene-text reading, and visual question answering~\cite{lin2014microsoft,antol2015vqa,singh2019towards,liu2024ocrbench} to broader suites that combine multimodal perception, knowledge, and reasoning~\cite{liu2024mmbench,yu2023mm,yue2024mmmu}.
\textit{Robustness benchmarks} test whether such capabilities persist under challenging inputs, either by corrupting images while preserving the reference label~\cite{hendrycks2019benchmarking}, or by using misleading questions and natural adversarial samples to expose failures in visually grounded reasoning~\cite{liu2024seeing,li2024naturalbench}.
\textit{Hallucination diagnostics} examine whether model claims are actually supported by image evidence~\cite{li2023evaluating}. These benchmarks primarily assess answer reliability, but provide limited systematic evaluation of semantic target recognition from incomplete, transformed, or embedded visual cues, as shown in Figure~\ref{fig:teaser}.

\begin{table}[h]
\centering
\scriptsize
\setlength{\tabcolsep}{1.1pt}
\renewcommand{\arraystretch}{0.90}
\begin{threeparttable}
\caption{Comparison between representative benchmarks~\tnote{1}.}
\label{tab:benchmark-comparison}
\vspace{-0.5em}
\begin{tabularx}{\columnwidth}{@{}>{\raggedright\arraybackslash}p{0.25\columnwidth}>{\raggedright\arraybackslash}X*{4}{>{\centering\arraybackslash}p{0.055\columnwidth}}@{}}
\toprule
Benchmark & Primary focus & A~\tnote{2} & T & C & H \\
\midrule
COCO Caption & Explicit-image captioning & -- & -- & -- & -- \\
VQAv2 & Visible-evidence VQA & -- & -- & -- & -- \\
TextVQA & Scene-text VQA & -- & -- & -- & -- \\
OCRBench & OCR perception & -- & -- & -- & -- \\
MMBench & Broad multimodal skills & -- & -- & -- & -- \\
MM-Vet & Integrated capabilities & -- & -- & -- & -- \\
MMMU & Expert multimodal reasoning & -- & -- & -- & \ding{51} \\
POPE & Hallucination diagnosis & -- & -- & -- & -- \\
ImageNet-C & Low-level corruption & P & -- & \ding{51} & -- \\
NaturalBench & Natural adversarial VQA & -- & -- & -- & \ding{51} \\
PQA & Gestalt grouping tasks & -- & -- & -- & \ding{51} \\
IllusionBench & Illusion understanding & \ding{51} & P & P & \ding{51} \\
\midrule
\textbf{GestaltBench} & \textbf{Hidden-target recovery} & \ding{51} & \ding{51} & \ding{51} & \ding{51} \\
\bottomrule
\end{tabularx}  
\begin{tablenotes}
    \item[1] A/T/C/H: designed ambiguity, hidden-target recovery, matched control, and measured human reference.
    \item[2] \ding{51}: yes; P: partial or subset-level; --: no.
\end{tablenotes}
\end{threeparttable}
\vspace{-0.5em}
\end{table}

\begin{figure*}[htbp]
\centering
\includegraphics[width=0.95\textwidth]{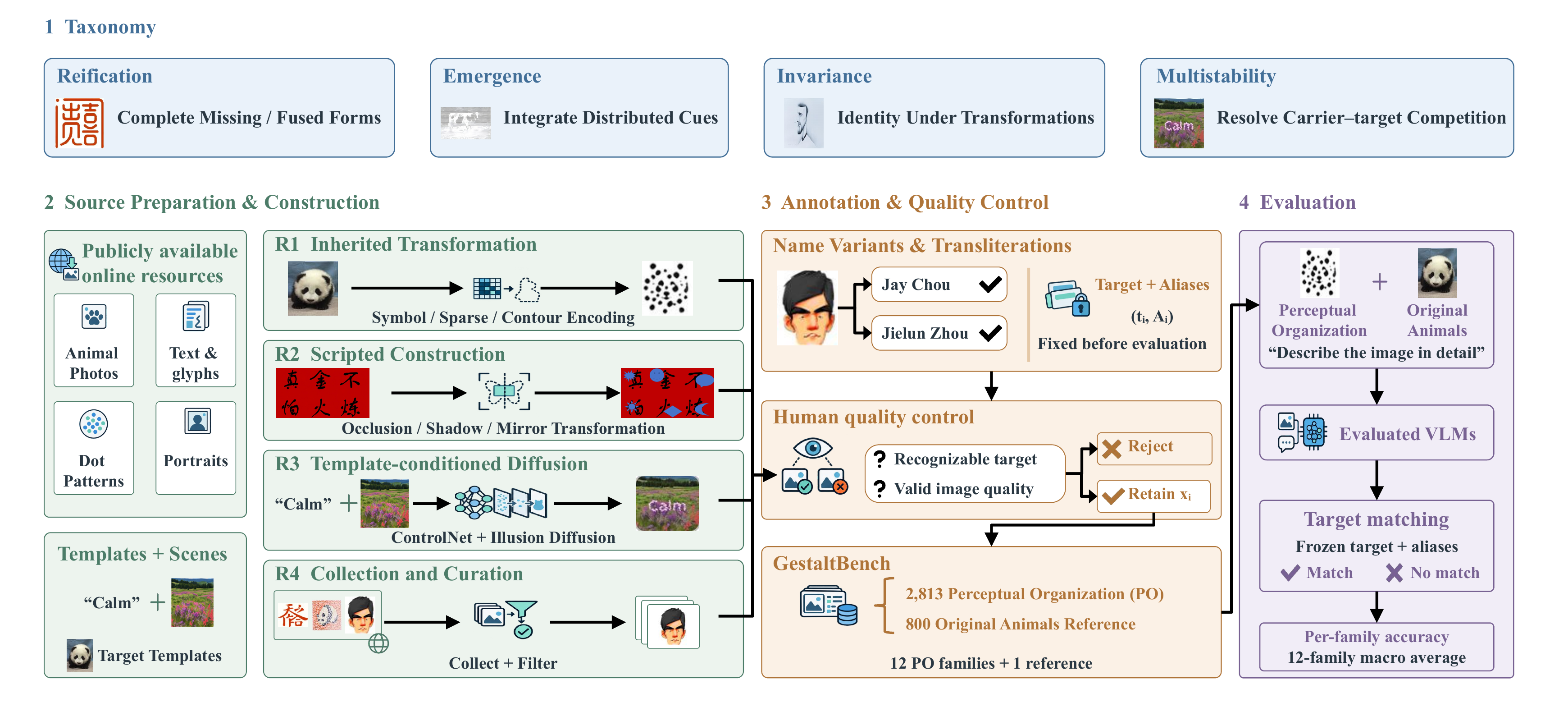}
\caption{GestaltBench construction and evaluation pipeline. (1) Twelve perceptual organization (PO) task families are organized into four operational Gestalt categories: Reification, Emergence, Invariance, and Multistability. (2) Candidates are generated or curated through four construction routes. (3) Human screening checks target recognizability and image quality; targets and accepted aliases are frozen before evaluation. The benchmark contains 2,813 PO images and 800 Original Animals (OA) reference images. (4) Evaluation: VLMs generate responses under the same open-ended prompt, scored by target matching.}
\label{fig:benchmark-overview}
\end{figure*}

\textbf{Perceptual Organization.}
Perceptual organization (PO) refers to the process of grouping local visual elements into coherent global percepts, and prior benchmarks have probed it from different angles. PQA builds synthetic tasks around Gestalt grouping laws, requiring models to infer a rule from an example and generate an answer grid~\cite{qi2021pqa}. A second line of work uses visual illusions: IllusionBench covers classical and real-scene illusions, Ishihara patterns, and trap illusions, evaluating both question answering and open-ended description~\cite{zhang2025illusionbench}, while IllusionVQA assesses illusion comprehension and localization through multiple-choice questions, including hidden illusions~\cite{shahgir2024illusionvqachallengingopticalillusion}.
These benchmarks cover selected aspects of PO tasks, including rule induction over synthetic grids and illusion perception, while PO tasks spanning a broader range of families remain to be explored.
Beyond coverage, a benchmark suited to diagnosing holistic perceptual organization should combine four design components: designed ambiguity, explicit hidden-target recognition, matched control sets, and measured human references. As summarized in Table~\ref{tab:benchmark-comparison}, existing benchmarks cover only a subset of these components.

\section{GestaltBench: Design and Construction}

GestaltBench evaluates holistic perceptual organization in VLMs: the ability to assemble local visual cues into coherent global semantic percepts. Figure~\ref {fig:benchmark-overview} summarizes the construction and evaluation pipeline.
The benchmark design follows four principles. First, the local cues in the constructed images are incomplete, transformed, or distributed so that they do not reveal it on their own. Second, targets are human-recognizable, verified by manual screening and a sampled human study. Next, the benchmark covers diverse Gestalt-based categories and target types. Finally, each image has a designated target with accepted aliases, and only target recognition earns credit.

We represent image construction as $s_i \rightarrow c_i \rightarrow x_i$, where $s_i$ is the source material, $c_i$ is a candidate image generated or collected from that source, and $x_i$ is the final image retained after quality screening. A source may yield multiple candidates; for collection and curation, the source image itself can serve as $c_i$.
Each final image has an answer annotation $(t_i, \mathcal{A}_i)$: $t_i$ is the semantic target (e.g., the word, phrase, or person identity), and $\mathcal{A}_i$ contains its accepted aliases, such as synonyms or alternative names. The VLM receives $x_i$ and is scored against $\{t_i\}\cup\mathcal{A}_i$ under the family-specific matching rules.
We group instances into task families and assign each family a category according to its recognition operation property.

\subsection{Gestalt-Based Operational Taxonomy}
\label{sec:gestalt-mapping}

\begin{table*}[t]
\centering
\caption{Gestalt-based operational taxonomy and construction design.
Item counts are family totals; Original Animals (OA) is the reference set outside the PO families.}
\label{tab:design-matrix}
\vspace{-0.5em}
\begin{threeparttable}
\renewcommand{\arraystretch}{0.92}\resizebox{\textwidth}{!}{
\begin{tabular}{@{}lllllc@{}}
\toprule
\textbf{Categories} &
\textbf{Family} &
\textbf{Target Type} &
\textbf{Recognition Demand} &
\textbf{Construction Operation} &
\textbf{Route}~\tnote{1} \\

\midrule

\multirow{3}{*}{Reification (243)}
& R-FSPO
& Rendered phrase
& Restore removed structural components
& Stroke or component occlusion
& R2
\\

& R-CCE
& Chinese idiom (single fused glyph)
& Separate fused components into a complete form
& Curated composite-character images (easy)
& R4
\\

& R-CCH
& Chinese idiom (single fused glyph)
& Reconstruct heavily fragmented forms
& Curated composite-character images (hard)
& R4
\\

\midrule

\multirow{4}{*}{Emergence (944)}
& E-ACT
& Animal category
& Read a global shape from distributed symbols
& ASCII-symbol re-encoding of an OA target
& R1
\\

& E-BWSE
& Animal category
& Bind sparse contour fragments into one whole
& Black--white sparse contour encoding
& R1
\\

& E-CBS
& Digit, letter, word, or object
& Group chromatically to reveal a figure in a dot field
& Curated Ishihara-style dot images
& R4
\\

& E-DSV
& Chinese idiom or English phrase
& Integrate foreground and cast shadow into one target
& Foreground--shadow composition
& R2
\\

\midrule

\multirow{2}{*}{Invariance (542)}
& I-CCM
& Chinese phrase or English sentence
& Preserve text identity under reflection
& Mirrored composition
& R2
\\

& I-AP
& Person identity
& Preserve identity under abstraction
& Curated abstract portraits
& R4
\\

\midrule

\multirow{3}{*}{Multistability (1{,}084)}
& M-THW
& English word
& Resolve carrier--target competition at word level
& Mask-conditioned scene synthesis
& R3
\\

& M-THP
& English phrase
& Resolve carrier--target competition at phrase level
& Mask-conditioned scene synthesis
& R3
\\

& M-HAI
& Animal category
& Resolve carrier--target competition for an embedded object
& Animal-template-conditioned scene synthesis
& R3
\\

\midrule

Reference
& OA
& Animal category
& Direct animal recognition
& Quality-scored photograph selection
& --
\\

\bottomrule
\end{tabular}
}
\begin{tablenotes}
\scriptsize
\item[1] Route represents the four construction routes mentioned in Section~\ref{sec:construction}.
\end{tablenotes}
\end{threeparttable}
\vspace{-0.5em}
\end{table*}

Gestalt psychology studies how relationships among local elements give rise to organized wholes, holding that the whole is more than the sum of its parts~\cite{wertheimer1923,wagemans2012century}. Dots arranged along a circular path, for example, are perceived as a circle.

Classic Gestalt grouping laws overlap and vary across accounts, so they do not offer clean, mutually exclusive categories for a benchmark.
Instead, we organize the benchmark around four higher-level holistic categories~\cite{lehar2003gestalt,hushagen2021depth}. \emph{Reification} supplies form absent from the immediate evidence, here by completing occluded or fused components into a coherent textual form. \emph{Emergence} yields a global whole from distributed cues, with the target recognized through the overall spatial or chromatic structure of seemingly disordered local elements. \emph{Invariance} preserves text or person identity across transformations such as reflection and abstraction. \emph{Multistability} permits competing stable interpretations, instantiated here by a coherent carrier scene and an embedded target in the same image.

These categories describe the operations needed to recognize a target and form our operational categories. Each family is assigned the category that best describes its dominant recognition operation.
\textbf{Reification:} This category covers two reconstruction conditions: missing components under occlusion and character structures obscured by fusion. Font Secondary Parts Occlusion (R-FSPO) calls for inferring occluded strokes or components from the remaining textual structure. Separating fused components and reconstructing complete character forms are central to Composite Characters Easy (R-CCE) and Composite Characters Hard (R-CCH), with more complex fusion, fragmentation, and deformation in the latter.
\textbf{Emergence:} Different cue organizations can yield a coherent whole, from symbolic and contour patterns to chromatic and foreground--shadow groupings. In ASCII Code Transformation (E-ACT) and Black-White Spot Enhancement (E-BWSE), discrete characters and sparse black--white contour fragments, respectively, must be integrated into global shapes. Chromatic grouping separates a figure from colored dots in Color Blindness Simulation (E-CBS), while recognition in Different Shadow Variations (E-DSV) depends on combining foreground characters and cast shadows into one textual interpretation.
\textbf{Invariance:} Identity preservation spans geometric reflection and artistic changes in depiction, covering recognition across altered textual forms and portrait appearances. Composite Characters Mirrored (I-CCM) probes textual identity when character forms are reflected. Recognizing stable identity cues through stylization, abstraction, and deformation is the corresponding demand in Abstract Portraits (I-AP).
\textbf{Multistability:} The same carrier--target competition takes different visual forms when the embedded structure is a word, a phrase, or an animal. Text Hiding Words (M-THW) and Text Hiding Phrases (M-THP) involve perceptually separating embedded word and phrase structures from coherent carrier scenes, extending recognition from individual words to longer textual units. Hidden Animal Images (M-HAI) extends the same separation operation to animal structures.
Together, these families provide diagnostic coverage of the four properties without claiming to exhaust their possible visual realizations.
Original Animals (OA) provides 800 reference images outside the categories. Table~\ref{tab:design-matrix} maps the categories to recognition mechanisms, construction routes, and targets.

\subsection{Source Preparation and Construction}
\label{sec:construction}

The family construction procedures below are organized by the four categories, covering source preparation and candidate generation. 
\textbf{Reification:} R-FSPO starts from publicly available text images and uses scripts to crop or mask secondary strokes or components, producing occluded candidates. R-CCE and R-CCH collect online composite-character images and annotate their constituent text. R-CCE selects conventional glyphs with simple cropping, splitting, or local composition, while R-CCH covers artistic glyphs with more complex fusion and deformation. 
\textbf{Emergence:} E-ACT and E-BWSE take OA reference photographs as their source. OA retains the top 800 of 1,740 photographs across 19 animal categories in the KTH animal-image collection (45.98\% retention). E-ACT then renders these photographs as symbol grids, while E-BWSE converts them into sparse black--white patterns; both use inherited transformation, retaining the animal labels and correspondence with OA. E-CBS collects existing Ishihara-style colored-dot images directly as candidates, and E-DSV prepares foreground--shadow compositions from publicly available text images through scripted construction. 
\textbf{Invariance:} I-CCM takes publicly available text images and mirrors them through scripted construction, with the original text as the target. I-AP selects stylized, abstract, or deformed portraits from online sources as candidates and pairs them with annotated person identities. 
\textbf{Multistability:} M-THW and M-THP select 20 English words and 114 English phrases, respectively, render them as text templates, and embed them in carrier scenes with template-conditioned diffusion. M-HAI selects 12 animal instances from six categories as templates and combines each with 16 scenes at three blending levels, generating 576 images through template-conditioned diffusion.

Overall, these construction procedures can be grouped into four routes for producing candidate images $c_i$ from source material $s_i$. \textbf{(R1) Inherited Transformation} re-encodes OA reference photographs while preserving the animal target $t_i$. Symbol encoding applies adaptive contrast stretching, gamma adjustment, and mild sharpening before rendering with a character atlas. Black--white encoding applies grayscale conversion, contrast enhancement, thresholding, and structured sparse deletion. \textbf{(R2) Scripted Construction} applies cropping or masking, foreground--shadow composition, or mirror transformation to source text images, preserving the complete text as $t_i$. \textbf{(R3) Template-conditioned Diffusion} uses ControlNet~\cite{zhang2023addingconditionalcontroltexttoimage} with Illusion Diffusion~\cite{burgert2023diffusionillusionshidingimages} to embed a target template within a carrier scene. Text is rendered as a binary mask; animal images are used directly as templates. The template and carrier scene form $s_i$, and diffusion generates $c_i$ with the word, phrase, or animal target $t_i$ embedded in the scene. \textbf{(R4) Collection and Curation} collects online images as candidates $c_i$ for manual cleaning and target annotation.

\subsection{Annotation and Quality Control}
For each retained image $x_i$, we annotate and standardize the answer pair $(t_i, \mathcal{A}_i)$. Animal aliases include common names, Chinese and English equivalents, and singular/plural variants. Text answers preserve the complete target string $t_i$, while person identities allow documented name variants and transliterations. We apply Unicode NFKC normalization and normalize case, surrounding whitespace, and specified punctuation and spacing differences. Targets $t_i$ and alias sets $\mathcal{A}_i$ are frozen before evaluation; no aliases are added based on VLM outputs.

Manual screening removes PO candidates $c_i$ whose targets $t_i$ are unrecognizable to humans or whose visual quality is compromised. Retained candidates become the final images $x_i$. Screening excludes all-black, all-white, and severely distorted E-ACT and E-BWSE candidates. I-AP candidates undergo cleaning and identity standardization. For M-THW, we discard six all-black candidates $c_i$ and their labels, leaving 394 final images $x_i$. The benchmark contains 2,813 PO images $x_i$ across 12 task families and 800 OA reference images, totaling 3,613 images. Table~\ref{tab:construction-retention} summarizes sources, candidate and final counts, and retention ratios.

\begin{table}[h]
\centering
\begin{minipage}{\columnwidth}
\centering
\caption{Data construction statistics: source items, candidate pool, final instances after screening, and retention rate.}
\label{tab:construction-retention}
\vspace{-0.5em}
\begin{threeparttable}
\fontsize{8pt}{9.5pt}\selectfont
\setlength{\tabcolsep}{1.5pt}
\renewcommand{\arraystretch}{0.92}
\begin{tabularx}{\linewidth}{@{}l>{\raggedright\arraybackslash}Xccc@{}}
\toprule
Family & Source & Can.~\tnote{1} & Final & \shortstack{Ret.(\%)~\tnote{2}} \\
\midrule
R-FSPO & 105 text images & 105 & 105 & 100.00 \\
R-CCE & Online composite-character images (easy) & 117 & 99 & 84.62 \\
R-CCH & Online composite-character images (hard) & 40 & 39 & 97.50 \\
E-ACT & 800 OA reference images & 800 & 321 & 40.13 \\
E-BWSE & 800 OA reference images & 800 & 306 & 38.25 \\
E-CBS & Online Ishihara-style images & 118 & 117 & 99.15 \\
E-DSV & 100 Chinese + 100 English phrase images & 200 & 200 & 100.00 \\
I-CCM & 107 Chinese + 107 English text images & 214 & 214 & 100.00 \\
I-AP & Online abstract portraits & 328 & 328 & 100.00 \\
M-THW & 20 English words & 400 & 394 & 98.50 \\
M-THP & 114 English phrases & 114 & 114 & 100.00 \\
M-HAI & 12 animal templates $\times$ 16 scenes & 960 & 576 & 60.00 \\
OA & KTH animal photographs & 1,740 & 800 & 45.98 \\
\midrule
Total & & 5936 & 3,613 & 60.87 \\
\bottomrule
\end{tabularx}
\begin{tablenotes}[flushleft]
\scriptsize
\item[1] Candidate and Final count images before and after screening; $c_i$ and $x_i$ denote these stages for PO families.
\item[2] Retention: percentage of candidates retained in the final dataset (Final/Candidate $\times$ 100\%).
\end{tablenotes}
\end{threeparttable}
\vspace{-0.5em}
\end{minipage}
\end{table}

\subsection{Evaluation Protocols and Metrics}
\label{sec:evaluation-metrics}

\textbf{Evaluation Protocols.} 
For each PO image $x_i$ in the 12 families, the VLM produces a description $y_i$ without task-specific hints. We normalize case, spacing, and punctuation in $y_i$, then match it against $t_i$ and $\mathcal{A}_i$ to determine whether the target is recognized. Primary scoring uses complete-string matching with family-specific supplementary rules. For R-FSPO, R-CCE/R-CCH, and Chinese I-CCM, all target characters must appear, regardless of order. Chinese E-DSV requires all foreground and shadow characters. M-THW/M-THP require explicit candidate text, numeric E-CBS answers require a pattern-recognition context, and I-AP requires name-boundary matching. Partial target recognition receives no credit.

\textbf{Evaluation Metrics.} 
We report primary target-recognition scores and separate auxiliary diagnostics.
\textbf{(1) Primary:} For response $y_i$ with canonical target $t_i$ and alias set $\mathcal{A}_i$, target recognition is $r_i=\mathbf{1}\!\left[\operatorname{match}\!\left(\operatorname{norm}(y_i),\{t_i\}\cup\mathcal{A}_i\right)\right]$. The recognition rate for family $f$ with $N_f$ items is $S_f=\frac{100}{N_f}\sum_{i=1}^{N_f}r_i$. Each $S_f$ reports the percentage of items with complete target recognition, giving the score the same interpretation across families. We summarize performance on PO images with the equal-weight macro-average $S_{\mathrm{macro}}=\frac{1}{12}\sum_f S_f$, which prevents the largest family from dominating. This aggregate summarizes recognition across distinct mechanisms and target structures without defining a universal ranking.
\textbf{(2) Auxiliary:} We report auxiliary scores only for E-ACT, E-BWSE, and I-CCM to examine residual information when complete target recognition fails. For E-ACT and E-BWSE, paired-description consistency (PDC) averages token-set Jaccard similarity, word-frequency cosine similarity, and normalized character-level Levenshtein similarity between the response $y_i$ to the final PO image $x_i$ and the same model's response to its OA reference source $s_i$. For I-CCM's Chinese subset, Char-F1 summarizes character-set precision and recall between the full response $y_i$ and canonical target $t_i$, ignoring order and repetition. These diagnostics are excluded from the 12-family macro-average.

\section{Experimental Setup}

\textbf{Settings.}
All models are queried once per item in an open-ended image-to-text setting without task-specific tuning. Each VLM is asked to describe the image in detail, with a family-specific hint that names only the expected target type (e.g., \emph{animal} for M-HAI).
The generation temperature is set to 0.2, and each response is limited to 1,024 tokens. Nucleus (top-$p$) and top-$k$ sampling are left at each endpoint's default. Each item generates one free-form response, without demonstrations, retries, resampling, voting, or tools. Responses are scored with the established family-specific protocols and metrics. The detailed settings are listed in Appendix~\ref{app:exp-details}.

\textbf{Models.}
We evaluate both API-based and open-weight models drawn from several model families. \textbf{(1) API-based models:} the GPT family includes GPT-4o, GPT-5.1, and GPT-5.5~\cite{openai2026models}; the Gemini family includes Gemini-2.5-Flash, Gemini-2.5-Pro, and Gemini-3.5-Flash~\cite{google2026models}; Qwen3.7-Plus~\cite{alibaba2026qwen37plus} and GLM-5V-Turbo~\cite{zai2026glm5vturbo}; \textbf{(2) Open-weight models:} Llama-4-Maverick~\cite{meta2025llama4}, LLaVA-Next-110B~\cite{li2024llavanextstronger}, and Mistral-Large-3~\cite{mistral2025mistrallarge3}.

\textbf{Research Questions.}
The experiments are organized to answer the following five research questions:
\begin{enumerate}
    \item \textbf{Recognition Performance:} How well do VLMs recognize semantic targets in perceptual organization tasks, relative to the reference group (OA set)?
    \item \textbf{Information under Failure:} Is target-recognition failure all-or-nothing, or do models retain partial evidence?
    \item \textbf{Across-Task Consistency:} Is the model's recognition performance consistent across perceptual organization task families, or tied to particular recognition mechanisms?
    \item \textbf{Model-Version Variation:} Do newer VLM releases within the same provider improve target recognition?
    \item \textbf{Human Reference:} Can humans recognize the hidden targets, and how do VLMs compare under the same protocol?
\end{enumerate}

\section{Experimental Results}

\begingroup
\setlength{\fboxsep}{0.45pt}
\definecolor{blue}{HTML}{367DBD}
\definecolor{orange}{HTML}{F4543C}
\definecolor{green}{HTML}{00BCC9}
\newcommand{\bestscore}[1]{\colorbox{black!18}{\strut\textbf{#1}}}
\newcommand{\secondscore}[1]{\colorbox{black!7}{\strut\underline{#1}}}
\newcommand{\scorechange}[2]{\makebox[0pt][l]{\hspace{2pt}$_{\textcolor{#1}{#2}}$}}
\newcommand{\meanone}[2]{\makebox[\linewidth][c]{\makebox[23pt][c]{#1}\makebox[18pt][l]{#2}}}
\begin{table*}[t]
\centering
\caption{Binary target-recognition accuracy (percent) for 11 VLMs and the seven-participant human reference study.}
\label{tab:main-results}
\vspace{-0.5em}
\small
\setlength{\tabcolsep}{1pt}
\renewcommand{\arraystretch}{1.05}
\captionsetup{font=footnotesize,skip=2pt}
\begin{threeparttable}
\begin{tabularx}{\linewidth}{@{}>{\raggedright\arraybackslash}p{0.16\linewidth}*{13}{>{\centering\arraybackslash}X}>{\centering\arraybackslash}p{0.082\linewidth}>{\centering\arraybackslash}p{0.077\linewidth}@{}}
\toprule
\multicolumn{1}{c}{\multirow{2}{*}{\textbf{Model}}} & \multicolumn{3}{c}{\textbf{Reification}} & \multicolumn{4}{c}{\textbf{Emergence}} & \multicolumn{2}{c}{\textbf{Invariance}} & \multicolumn{3}{c}{\shortstack{\textbf{Multistability}}} & \multicolumn{2}{c}{\shortstack{\textbf{Reference Set}}} & \multirow{2}{*}{\textbf{Mean}$_{\mathbf{12}}$~\tnote{1}} \\
\cmidrule(lr){2-4}\cmidrule(lr){5-8}\cmidrule(lr){9-10}\cmidrule(lr){11-13}\cmidrule(lr){14-15}
 & FSPO & CCE & CCH & ACT & BWSE & CBS & DSV & CCM & AP & THW & THP & HAI & OA & \shortstack{Mean}$_{\mathrm{AB}}$~\tnote{2} & \\
\midrule
\multicolumn{16}{l}{\textbf{API-Based Models}} \\[1pt]
\hspace*{8pt}GPT-4o & 0.00 & 0.00 & 0.00 & \bestscore{47.35}\makebox[0pt][l]{\tnote{3}} & 83.01 & \secondscore{45.69} & 18.50 & 50.00 & 0.00 & 4.57 & 7.89 & 1.56 & 79.38 & \meanone{\bestscore{65.18}}{\scorechange{green}{14.2\,\downarrow}} & 21.55  \\
\hspace*{8pt}GPT-5.1 & 0.00 & 0.00 & 0.00 & \secondscore{44.55} & \secondscore{85.29} & \bestscore{46.55} & 23.50 & 50.00 & 0.00 & 4.31 & 11.40 & 1.39 & 78.63 & \meanone{\secondscore{64.92}}{\scorechange{green}{13.7\,\downarrow}} & 22.25  \\
\hspace*{8pt}GPT-5.5 & 0.95 & 0.00 & 0.00 & 0.93 & 72.88 & 26.72 & 28.00 & 49.07 & 3.96 & 2.03 & 0.88 & 1.39 & \secondscore{79.88} & \meanone{36.91}{\scorechange{green}{43.0\,\downarrow}} & 15.57  \\
\hspace*{8pt}Gemini-2.5-Flash & 40.95 & 0.00 & 0.00 & 4.67 & 79.41 & 5.13 & 33.50 & 34.11 & 4.27 & 6.35 & \secondscore{16.67} & 2.95 & 78.88 & \meanone{42.04}{\scorechange{green}{36.8\,\downarrow}} & 19.00  \\
\hspace*{8pt}Gemini-2.5-Pro & 60.00 & \bestscore{1.01} & 2.56 & 4.67 & 80.07 & 9.40 & 24.50 & 0.00 & 4.88 & \secondscore{7.11} & 15.79 & \bestscore{5.03} & 79.63 & \meanone{42.37}{\scorechange{green}{37.3\,\downarrow}} & 17.92  \\
\hspace*{8pt}Gemini-3.5-Flash & \bestscore{78.10} & \bestscore{1.01} & \bestscore{7.69} & 0.00 & 82.35 & 39.32 & 20.50 & 0.93 & \bestscore{7.93} & \bestscore{8.38} & 14.04 & 2.95 & 77.75 & \meanone{41.18}{\scorechange{green}{36.6\,\downarrow}} & 21.93  \\
\hspace*{8pt}Qwen3.7-Plus & \secondscore{74.29} & 0.00 & \secondscore{2.94} & 0.00 & 77.78 & 26.50 & \bestscore{78.50} & \secondscore{50.70} & 6.10 & 4.57 & 5.26 & 0.69 & 79.25 & \meanone{38.89}{\scorechange{green}{40.4\,\downarrow}} & \secondscore{27.28}  \\
\hspace*{8pt}GLM-5V-Turbo & 60.00 & 0.00 & 0.00 & 0.00 & \bestscore{88.24} & 39.66 & \secondscore{73.00} & \bestscore{78.50} & \secondscore{7.01} & 4.31 & 2.63 & 2.95 & \bestscore{81.38} & \meanone{44.12}{\scorechange{green}{37.3\,\downarrow}} & \bestscore{29.69}  \\
\midrule
\multicolumn{16}{l}{\textbf{Open-Weight Models}} \\[1pt]
\hspace*{8pt}Llama-4-Maverick & 0.00 & 0.00 & 0.00 & 20.56 & 70.59 & 11.97 & 12.50 & 3.74 & 1.22 & 3.30 & \bestscore{17.54} & 1.56 & 76.25 & \meanone{45.58}{\scorechange{green}{30.7\,\downarrow}} & 11.91  \\
\hspace*{8pt}LLaVA-Next-110B & 0.00 & 0.00 & 0.00 & 36.45 & 68.95 & 3.42 & 8.00 & 0.00 & 0.00 & 4.06 & \bestscore{17.54} & 2.43 & 74.38 & \meanone{52.70}{\scorechange{green}{21.7\,\downarrow}} & 11.74  \\
\hspace*{8pt}Mistral-Large-3 & 0.00 & 0.00 & 0.00 & 0.00 & 53.59 & 8.55 & 10.50 & 2.80 & 0.00 & 1.27 & 0.00 & \secondscore{3.13} & 76.88 & \meanone{26.80}{\scorechange{green}{50.1\,\downarrow}} & 6.65  \\
\midrule
\hspace*{2pt}Human mean (\emph{n}\,=\,7) & 62.86 & 62.86 & 42.86 & 68.57 & 71.43 & 65.71 & 77.14 & 37.14 & 57.14 & 82.86 & 88.57 & 80.00 & -- & \meanone{70.00}{} & 66.43 \\
\bottomrule
\end{tabularx}
\begin{tablenotes}
\scriptsize
\item[1] Mean$_{12}$ indicates the average recognition rate of the 12 familyies; Mean$_{\mathrm{AB}}$ indicates the average recognition rate of E-ACT and E-BWSE.
\item[2] \shortstack{Mean}$_{\mathrm{AB}}$ subscripts show percentage-point changes from the OA recognition rate. \textcolor{orange}{$\uparrow$} denotes an increase, \textcolor{green}{$\downarrow$} a decrease, and \textcolor{blue}{$0.0$} a change rounding to zero.
\item[3] Bold and underline mark the best and second-best distinct VLM scores in each column; ties receive the same emphasis.
\end{tablenotes}
\end{threeparttable}
\vspace{-0.5em}
\end{table*}
\endgroup

\subsection{RQ1: Recognition Performance}
\label{sec:experiments-control}

Table~\ref{tab:main-results} reports the binary target-recognition rate for all 11 VLMs. Across the 12 PO task families, $\mathrm{Mean}_{12}$ ranges from 6.65\% to 29.69\%; even the strongest model remains below 30\%. GLM-5V-Turbo achieves the highest score at 29.69\%, followed by Qwen3.7-Plus at 27.28\% and GPT-5.1 at 22.25\%. The evaluated I-API-based models score 15.57--29.69\%, exceeding the 6.65--11.91\% range of the evaluated open-weight models. 
For reference group OA, the recognition rate is substantially higher, at 74.38--81.38\%, providing a qualitative baseline. Compared with OA, the 12 PO subsets yield markedly lower recognition rates across models, suggesting that target recognition under designed tasks remains a challenge for the VLMs’ holistic perceptual organization abilities.

The overall averages conceal substantial differences in task difficulty. All models exceed 50\% on E-BWSE, whereas the highest model score remains below 20\% in the other six families: R-CCE, R-CCH, I-AP, M-THW, M-THP, and M-HAI. 
R-CCE is particularly difficult, with a best score of only 1.01\% and zero recognition for nine of the eleven models. Even on M-THP, the family with the highest peak among these six, the best score is only 17.54\%. Although VLMs also show limitations in the other six families, their recognition performance is particularly poor on these six.

Comparison between OA and E-ACT \& E-BWSE provides a direct and matched result, since E-ACT and E-BWSE re-encode animal targets drawn from the OA source pool. 
Their average recognition rate, $\mathrm{Mean}_{\mathrm{AB}}$, ranges from 26.80\% to 65.18\%, falling below each model's OA recognition rate by 13.7--50.1 percentage points. Separating the two constructions reveals markedly different outcomes: E-ACT recognition is 0--47.35\%, while E-BWSE recognition is 53.59--88.24\%, with E-BWSE outperforming E-ACT for every model. Six models even score higher on E-BWSE than on OA; GLM-5V-Turbo, for example, reaches 88.24\% on E-BWSE versus 81.38\% on OA. A plausible explanation is that E-BWSE preserves the animal's global shape in sparse black--white contours while suppressing local texture, which may make shape cues easier to isolate. E-ACT instead encodes animal form through the spatial density of an ASCII-like symbol grid, requiring the model to integrate discrete characters into a global shape. This difference may explain why recognition is generally stronger from E-BWSE contours than from E-ACT symbol patterns.

\textbf{Key Findings.} 
Designed tasks pose a substantial challenge to the holistic perceptual organization capability of current VLMs. Averaged over the 12 PO families, recognition is only 18.68\%, far below that on reference images. On the matched comparison, recognition falls from 78.39\% on OA to 45.52\% on E-ACT \& E-BWSE (41.90\% decrease), which are re-encoded from OA.

\subsection{RQ2: Information under Failure}
\label{sec:auxiliary-results}

\begingroup
\begin{table}[!ht]
\centering
\caption{Auxiliary scores beyond target recovery: description consistency and target-character overlap.}
\label{tab:auxiliary-diagnostics}
\vspace{-0.5em}
\begin{threeparttable}
\small
\setlength{\tabcolsep}{5.0pt}
\renewcommand{\arraystretch}{0.84}
\captionsetup{font=footnotesize,skip=2pt}
\begin{tabularx}{\linewidth}{@{}>{\raggedright\arraybackslash}Xccc@{}}
\toprule
Model & E-ACT~\tnote{1} & E-BWSE & I-CCM~\tnote{2} \\
\midrule
GPT-4o & 0.359 & 0.413 & 0.047 \\
GPT-5.1 & 0.354 & 0.387 & 0.105 \\
GPT-5.5 & 0.283 & 0.352 & 0.089 \\
Gemini-2.5-Flash & 0.362 & 0.434 & 0.235 \\
Gemini-2.5-Pro & \textbf{0.387} & \underline{0.443} & 0.252 \\
Gemini-3.5-Flash & 0.255 & 0.334 & 0.126 \\
Qwen3.7-Plus & 0.255 & 0.367 & \underline{0.351} \\
GLM-5V-Turbo & 0.204 & 0.334 & \textbf{0.931} \\
Llama-4-Maverick & 0.370 & \textbf{0.448} & 0.036 \\
LLaVA-Next-110B & \underline{0.375} & 0.420 & 0.033 \\
Mistral-Large-3 & 0.305 & 0.406 & 0.150 \\
\midrule
Mean & 0.319 & 0.394 & 0.214 \\
\bottomrule
\end{tabularx}
\begin{tablenotes}
    \scriptsize
    \item[1] The auxiliary score of E-ACT and E-BWSE is PDC; PDC $=(J+C+L)/3$ compares each response with the same model's matched-OA response. 
    \item[2] I-CCM Char-F1 is an order-free, whole-response exploratory upper bound on incomplete reconstruction, computed on the Chinese-script subset.
\end{tablenotes}
\end{threeparttable}
\vspace{-0.5em}
\end{table}
\endgroup

Table~\ref{tab:auxiliary-diagnostics} uses two complementary diagnostics to characterize information beyond primary recognition. Char-F1 measures overlap with target characters on the Chinese-script subset of I-CCM, while PDC measures consistency between descriptions of E-ACT/E-BWSE images and their matched OA sources. Together, they reveal character overlap and shared descriptive content that binary recognition rate alone does not capture.

Across models, I-CCM Char-F1 averages 0.214; GLM-5V-Turbo scores highest at 0.931, followed by Qwen3.7-Plus at 0.351. Gemini-2.5-Pro obtains 0.252 despite recognizing no complete I-CCM targets, showing that target characters can still appear in unsuccessful responses. All models have higher PDC on E-BWSE than on E-ACT (model means: 0.394 versus 0.319), indicating greater consistency with their OA descriptions. Gemini-2.5-Pro leads E-ACT PDC at 0.387, while Llama-4-Maverick leads E-BWSE PDC at 0.448. However, these auxiliary rankings differ from target-recognition rankings: GPT-4o leads E-ACT recognition at 47.35\%, and GLM-5V-Turbo leads E-BWSE recognition at 88.24\% despite a PDC of only 0.334. Char-F1 ignores character order and scans the full response, while PDC measures lexical similarity to a model-generated description. Both are exploratory diagnostics and are excluded from $\mathrm{Mean}_{12}$.

\textbf{Key Findings.}
Even when complete target recognition fails, models can still reproduce part of the target characters. Auxiliary metrics complement the recognition rate.

\subsection{RQ3: Across-Task Consistency}

\begin{figure}[htbp]
  \centering
  \vspace{-0.5em}
  \includegraphics[width=0.8\linewidth]{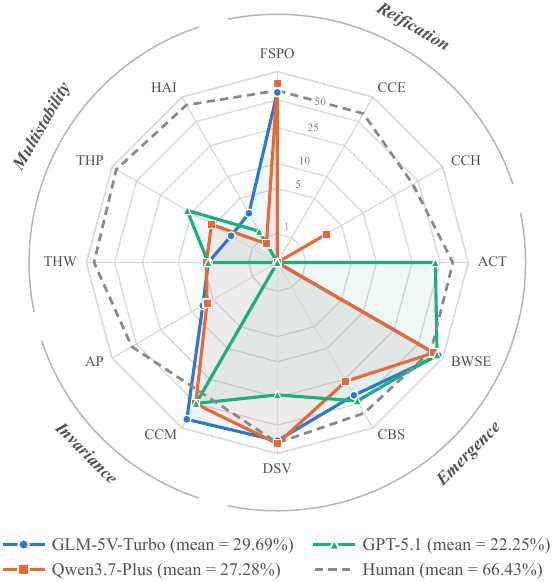}
  \caption{Per-family recognition rates of the three VLMs with the highest $\mathrm{Mean}_{12}$. The radius uses a $\log_{10}(1{+}x)$ scale to keep distinguishability; rings mark 1, 5, 10, 25, 50, and 100\%.}
  \label{fig:rq3-radar}
  \vspace{-0.5em}
\end{figure}

Table~\ref{tab:main-results} and Figure~\ref{fig:rq3-radar} together show that strong results concentrate in specific mechanisms: GLM-5V-Turbo reaches 88.24\% on E-BWSE, Gemini-3.5-Flash 78.10\% on R-FSPO, and Qwen3.7-Plus 78.50\% on E-DSV. I-CCM provides the clearest rank reversal together with a script split: GLM-5V-Turbo recognizes 168 of 214 complete targets (78.50\%) and is the only model above 2\% on the Chinese mirrored subset (67/107), whereas GPT-4o and GPT-5.1 transcribe all 107 mirrored English sentences yet recognize no Chinese item, and Gemini-2.5-Pro recognizes nothing in either script. E-ACT reverses the pattern again: GPT-4o recognizes 152 of 321 animal targets (47.35\%), whereas the macro-leading GLM-5V-Turbo recognizes none.

A model's lead is therefore tied to particular recognition operations rather than a stable global hierarchy: across the 66 family pairs, per-family model rankings agree only weakly (mean Spearman $\rho=0.164$, with 23 pairs non-positive and E-ACT \& I-AP lowest at $-0.690$). Category means are 36.26\% for Emergence, 16.15\% for Invariance, 9.98\% for Reification, and 5.64\% for Multistability-linked tasks. Yet the E-ACT \& E-BWSE spread within Emergence (14.47--76.56\%) is twice the spread between category means, indicating that family-level differences exceed those captured by the coarse taxonomy.
Figure~\ref{fig:rq3-radar} illustrates this for the three leading models: their profiles cross rather than nest, even the top models show no consistent ordering across families (all 11 models: details in Appendix~\ref{app:additional-results}).

Carrier-aware outputs still miss the target. M-HAI style match remains at 91.30--99.30\%, while animal recognition is only 0.69--5.03\%. I-AP descriptions reach 0.791 mean feature consistency, yet exact identity recognition peaks at 7.93\%. Full recognition peaks at 8.38\% on M-THW and 17.54\% on M-THP, and stays below 8\% on R-CCE and R-CCH. Thus, reporting text, scene style, or portrait attributes does not imply recognition of the intended lexical content, animal, or identity.

\textbf{Key Findings.}
Model advantages vary with the task family and its recognition demands. Every evaluated model has substantial weaknesses in some families, including models that lead the overall average or individual tasks.

\subsection{RQ4: Model-Version Variation}

\begin{figure}[htbp]
  \centering
  \vspace{-0.5em}
  \begin{subfigure}{0.48\linewidth}
    \centering
    \includegraphics[width=\linewidth]{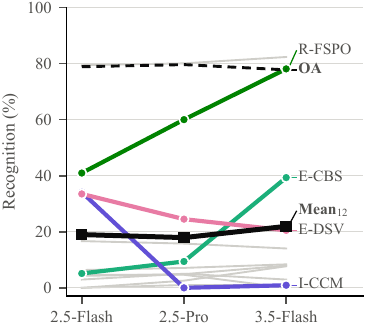}
    \caption{Gemini family}
    \label{fig:rq4-version-gemini}
  \end{subfigure}
  \hfill
  \begin{subfigure}{0.48\linewidth}
    \centering
    \includegraphics[width=\linewidth]{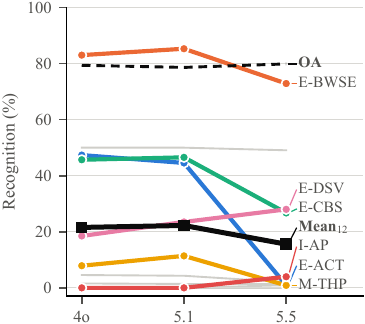}
    \caption{GPT family}
    \label{fig:rq4-version-gpt}
  \end{subfigure}
  \caption{Per-family recognition rate changes across model releases.}
  \label{fig:rq4-version}
\end{figure}

Figure~\ref{fig:rq4-version} traces the per-family changes behind Table~\ref{tab:main-results}: across releases, recognition performance does not improve consistently, suggesting that PO ability does not automatically improve alongside general model capabilities.

The Gemini variants (Figure~\ref{fig:rq4-version-gemini}) permit a descriptive comparison, though their labels do not isolate architecture, data, alignment, or serving configuration. Across the three variants, $\mathrm{Mean}_{12}$ is 19.00\% for 2.5-Flash, 17.92\% for 2.5-Pro, and 21.93\% for 3.5-Flash, while OA remains within 77.75--79.63\%. Because 2.5-Pro changes the model tier, the cleaner release comparison is Flash-to-Flash: Gemini-3.5-Flash gains 2.93 percentage points over Gemini-2.5-Flash, but the changes are uneven across families: R-FSPO ($+37.15$) and E-CBS ($+34.19$) improve, whereas I-CCM ($-33.18$) and E-DSV ($-13.00$) regress. Auxiliary scores show a similar lack of uniform progress: 2.5-Pro has the highest E-ACT PDC, E-BWSE PDC, and I-CCM Char-F1 within the Gemini group, and all three are lower for 3.5-Flash.

The GPT variants (Figure~\ref{fig:rq4-version-gpt}) show a small initial improvement followed by a larger decline: $\mathrm{Mean}_{12}$ is 21.55\% for GPT-4o, 22.25\% for GPT-5.1, and 15.57\% for GPT-5.5, while OA remains within 78.63--79.88\%. GPT-5.1 gains only 0.70 percentage points over GPT-4o; GPT-5.5 then loses 6.68 points relative to GPT-5.1 despite the highest GPT OA score (79.88\%). Relative to GPT-5.1, GPT-5.5's gains on E-DSV ($+4.50$) and I-AP ($+3.96$) do not offset losses on E-ACT ($-43.62$), E-BWSE ($-12.41$), E-CBS ($-19.83$), and M-THP ($-10.52$). Descriptive consistency also declines across the three releases: E-ACT PDC falls from 0.359 to 0.354 to 0.283, and E-BWSE PDC from 0.413 to 0.387 to 0.352. Thus, the newest evaluated GPT variant combines stronger recognition on the OA set with weaker PO recognition and lower paired-description consistency.

\textbf{Key Findings.}
Recognition performance does not improve consistently with model-version updates, and earlier releases can outperform newer variants from the same provider. These suggest that the PO ability measured by our benchmark should not be assumed to improve automatically alongside general model capabilities.

\subsection{RQ5: Human Reference}
\label{sec:human-study}

\begin{figure}[htbp]
  \centering
  \vspace{-0.5em}
  \includegraphics[width=0.8\linewidth]{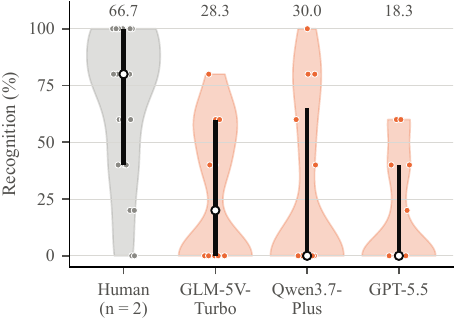}
  \caption{Comparison between humans and three VLMs on the same 60 sampled items.}
  \label{fig:rq5-violin}
  \vspace{-0.5em}
\end{figure}

Seven non-author participants each answered one of two counterbalanced 60-item forms, jointly covering 120 sampled items across the 12 families, as shown in Figure~\ref{fig:rq5-violin}. Questions specified the answer type without revealing the target; free-form responses were scored with the same alias-aware matching used for models. Human mean accuracy is 66.43\%; 107 of 120 items were solved by at least one participant and 80 by majority vote. The descriptive profiles vary by family (Table~\ref{tab:main-results}): the best VLM scores above the human mean on E-BWSE (88.24\% versus 71.43\%), whereas humans score higher on M-THP (88.57\% versus 17.54\%).
Across the 12 families, human scores are concentrated (median 67.14\%, no family below 37.14\%, 10 of 12 families at or above 50\%), whereas every VLM has a median of at most 8.16\% and at least six of the 12 families below 10\%; the long upper tails of the VLM violins are formed by a few families, mainly E-BWSE and, for the leading models, R-FSPO, E-DSV, and I-CCM.

To control for query format, three models completed the same 60-item form with the same questions and scoring as the human participants. GLM-5V-Turbo, Qwen3.7-Plus, and GPT-5.5 achieve 28.33\%, 30.00\%, and 18.33\%, respectively, compared with 66.67\% for the two humans assigned that form. The human--model difference thus persists under matched conditions. 
As shown in Figure~\ref{fig:rq5-violin}, the mean recognition rates of the three VLMs do not exceed 30\%, far below the human mean of 66.67\%, and their per-family scores are concentrated in the very low range (at or below 10.00\%).

\textbf{Key Findings.}
Humans recognize most sampled targets, with relative human and model performance varying by task family. Under matched questions and scoring, the tested VLMs remain below the human reference, while directed prompts improve detection but leave substantial recognition difficulties on the hardest tasks.

\section{Application: Toward Human Verification}
\label{sec:application}

Human verification, such as CAPTCHA, aims to distinguish human users from automated agents by posing challenges that are easy for humans but difficult for machines. The substantial Human--VLM performance gaps observed in the experiments on several PO families of GestaltBench suggest CAPTCHA design as a potential application. Motivated by this observation, we evaluate a selection-based CAPTCHA prototype informed by the benchmark results.

\subsection{From Target Recognition to CAPTCHA}

The benchmark results show contrasting Human--VLM recognition profiles across the four families considered for CAPTCHA construction. Human mean recognition reaches 82.86\% on M-THW and 80.00\% on M-HAI, exceeding the respective best VLM scores of 8.38\% and 5.03\% by 74.48 and 74.97 percentage points. E-ACT shows a smaller human advantage of 21.22 percentage points (68.57\% versus 47.35\%), whereas E-BWSE reverses the pattern: the human mean of 71.43\% is 16.81 percentage points below the best VLM score of 88.24\%.
The magnitude of these Human--VLM gaps arises from the PO tasks rather than incidental variation, suggesting that differences in PO target recognition could serve as a basis for human verification.

\begin{figure}[htbp]
\centering
\includegraphics[width=0.8\linewidth]{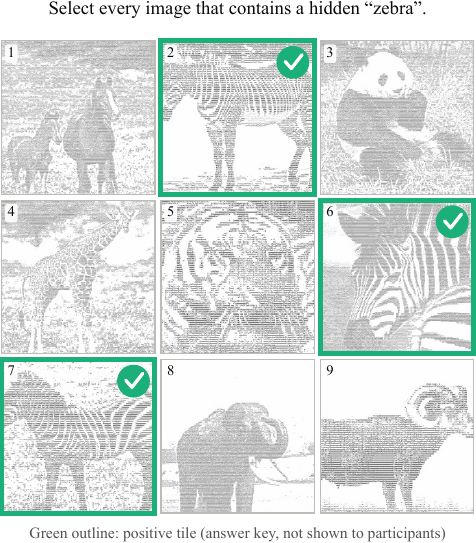}
\caption{An example of the selection-based CAPTCHA challenge (E-ACT). The participant is shown a $3{\times}3$ grid of nine image tiles and must select every tile that contains the specified target.}
\label{fig:application}
\end{figure}

We construct the selection-based CAPTCHA challenges using the four image families: E-BWSE, E-ACT, M-THW, and M-HAI. Each challenge presents a $3{\times}3$ grid of nine image tiles and asks the participant to select every tile matching a specified target, as shown in Figure~\ref{fig:application}. Two to four tiles are positives containing the target, while the remaining tiles are negatives. M-HAI images use the intermediate blending level. \emph{Exact-set accuracy} requires selecting all positives and no negatives; the chance level under uniformly random non-empty selection is $1/511$. For each family $f$, the Human--VLM gap $G_f$ is the human mean exact-set accuracy minus the three-model mean. Tile-level $F_1$ treats each of the nine tiles as a binary decision and reports the $F_1$ score of the positive class.

\subsection{Evaluation of the CAPTCHA Prototype}
\label{sec:challenge-study}

We evaluate 16 selection challenges, with four per family and a shared positive-count profile of 3, 2, 4, and 3 (48 positives across 144 tiles). Seven non-author users and three VLMs (GLM-5V-Turbo, Qwen3.7-Plus, and GPT-5.5) complete the same challenges as participants, with fixed family and tile order. The selection format, images, and target classes are held out from the benchmark, and the participants have not seen these tiles elsewhere.

\begin{figure}[htbp]
\centering
\includegraphics[width=\linewidth]{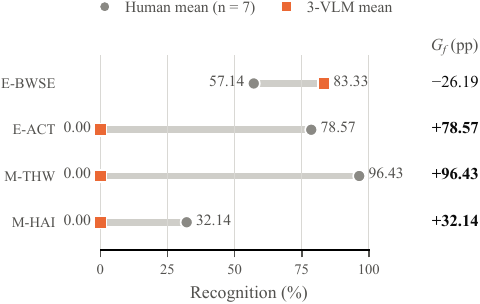}
\caption{Exact-set recognition rates on 16 held-out selection challenges (four per family). $G_f$ is the seven-human mean minus the three-VLM mean. Three VLMs are GLM-5V-Turbo, Qwen3.7-Plus, and GPT-5.5. Overall recognition rate is 66.07\% versus 20.83\% (45.24 pp); E-BWSE reverses because two models solve all four challenges.}
\label{fig:application-transfer}
\end{figure}

Humans achieve 66.07\% exact-set accuracy (74/112; tile-level $F_1=0.902$), compared with the three-model mean of 20.83\%, yielding an overall gap of 45.24 percentage points. GLM-5V-Turbo and Qwen3.7-Plus each achieve 25.00\%, while GPT-5.5 achieves 12.50\%; model tile-level $F_1$ ranges from 0.405 to 0.484. Human/model mean accuracy is 57.14/83.33\% on E-BWSE, 78.57/0\% on E-ACT, 96.43/0\% on M-THW, and 32.14/0\% on M-HAI, giving $G_f$ values of $-26.19$, 78.57, 96.43, and 32.14 percentage points, respectively, as shown in Figure~\ref{fig:application-transfer}. All model successes occur on E-BWSE, with GLM-5V-Turbo and Qwen3.7-Plus solving all four challenges. These results broadly align with the family-level Human--VLM performance patterns observed in the experiments.

These results provide proof-of-concept evidence that human advantages in target recognition can transfer to a selection-based verification task. M-THW and E-ACT emerge as promising carriers for distinguishing humans from the tested VLMs, while the model advantage on E-BWSE and low human success on M-HAI highlight the need to consider both model resistance and human solvability. GestaltBench can thus inform mechanism selection for CAPTCHA design, with prototype evaluation establishing whether the observed recognition profiles transfer to usable challenges. This workflow could support the development and validation of new verification challenges based on a broader range of perceptual mechanisms and interactive formats.

\section{Conclusion}
We introduce GestaltBench, a 3{,}613-item benchmark spanning 12 forms of perceptual organization (PO) task families and an 800-item original animal (OA) reference, to test recognition of intended semantic targets rather than carrier description. Across 11 VLMs and seven human participants, the best model reaches 29.69\% mean recognition versus 66.43\% for humans, despite 74--81\% model accuracy on controls. The gap persists under protocol-matched targeted questions, indicating mechanism-specific perceptual failures rather than prompt directness. A held-out CAPTCHA prototype produces a 45.24-point human--model gap; E-BWSE reverses as predicted, showing that family-level profiles can guide human-solvable, model-resistant challenge design.

Our findings are based on the task families and evaluation settings considered in this study, and are intended to be interpreted within this scope. Future work will extend the coverage of perceptual challenges and evaluation conditions, and further investigate the applicability of Gestalt-based human verification.

\section*{Ethics Statement}

GestaltBench is intended to support research on holistic perceptual organization in VLMs. The observed performance differences between human participants and the evaluated models motivate a proof-of-concept application to human verification. These findings are specific to the tested models, tasks, and evaluation conditions; they do not establish reliable separation of all human users from automated systems.

Human verification can introduce privacy risks when challenge responses, response times, or interaction logs are linked to identities or reused for profiling. Any deployment should minimize data collection and retention, limit data use to verification, and assess privacy risks before deployment. Reuse of publicly sourced images, particularly portraits, should also consider privacy and source attribution. Perceptual challenges may disadvantage users with visual impairments or different linguistic backgrounds, so accessible alternatives are necessary. Our prototype requires further assessment of privacy, accessibility, and resistance to adaptive attacks before operational use.


\bibliography{bib/references}
\bibliographystyle{arxiv}

\newpage
\appendix

\newpage
\appendix
\onecolumn

\providecommand{\authorcheck}[1]{\textbf{\textcolor{orange!85!black}{[TO CONFIRM: #1]}}}

\section{Additional Results}
\label{app:additional-results}

\begin{figure*}[h]

  \centering
  \includegraphics[width=0.92\textwidth]{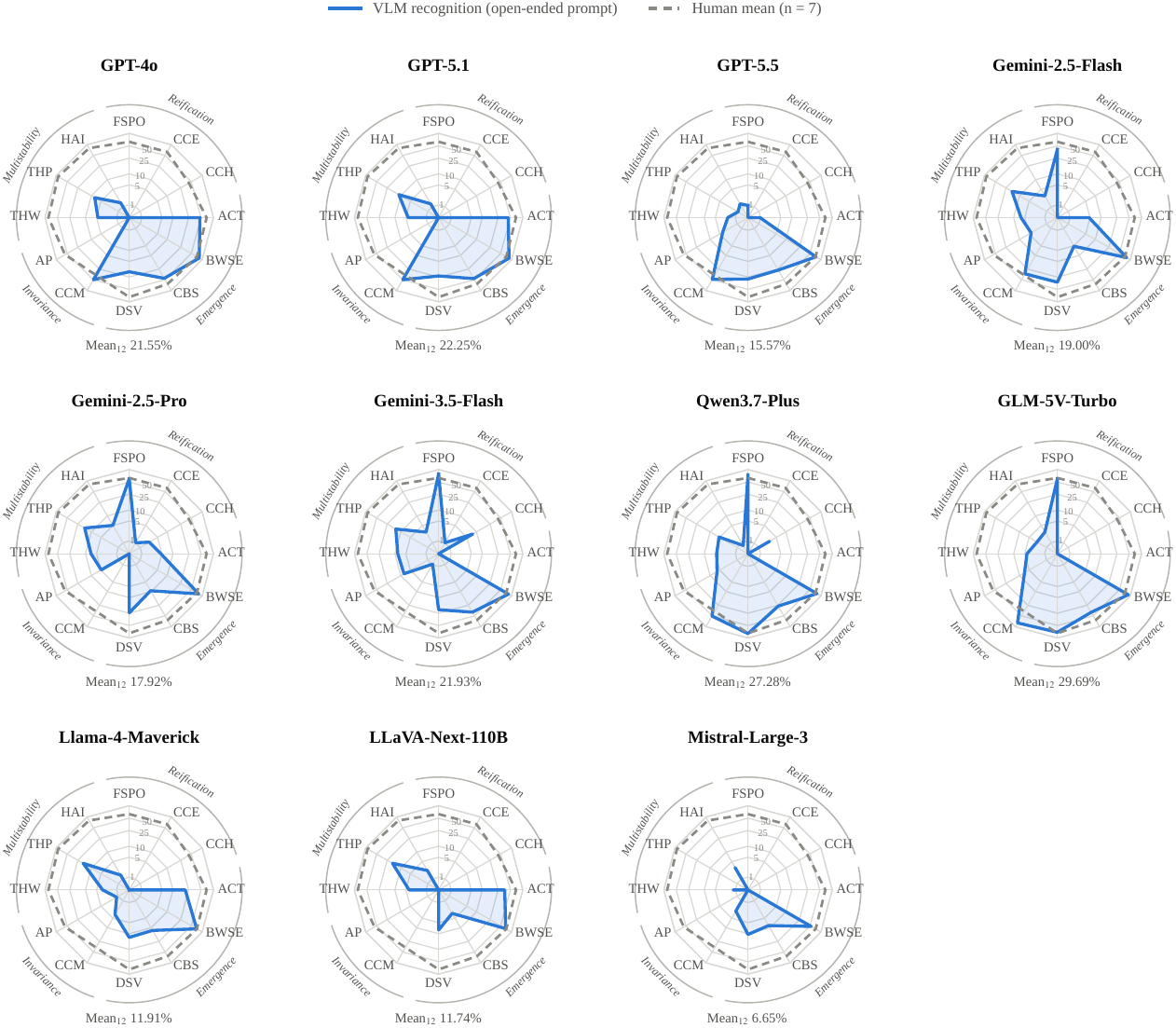}
  \caption{Per-family recognition profiles of all 11 VLMs (open-ended prompt; values from Table~\ref{tab:main-results}). Radius on a $\log_{10}(1{+}x)$ scale with rings at 1, 5, 10, 25, 50, and 100\%; dashed line: human mean ($n{=}7$).}
  \label{fig:app-radar-all}

\end{figure*}

Figure~\ref{fig:app-radar-all} extends Figure~\ref{fig:rq3-radar} to all 11 evaluated VLMs. All values are taken from Table~\ref{tab:main-results}; the radius uses the same $\log_{10}(1{+}x)$ scale, and the dashed line is the human mean of each family ($n{=}7$). The profiles confirm the observations of RQ3 and RQ4: GPT-4o, GPT-5.1, LLaVA-Next-110B, and Llama-4-Maverick are the only models above 5\% on E-ACT, R-FSPO is recognized almost exclusively by the Gemini, Qwen, and GLM models, E-DSV peaks for Qwen3.7-Plus and GLM-5V-Turbo, and no model approaches the human reference on R-CCE, R-CCH, I-AP, or the three Multistability families.

\section{Illustrative Examples}
\label{app:cases}

Table~\ref{tab:app-cases} shows two items with their image, a description of how the image is built, the frozen target and accepted alias set $\{t_i\}\cup\mathcal{A}_i$, and the verbatim responses of two models under the family-specific prompts of Table~\ref{tab:app-prompts} (long responses are abridged with [\dots]; nothing else is edited). The verdict is the output of the family-specific matching rule of Section~\ref{sec:evaluation-metrics}.

\begin{table*}[p]

\centering
\caption{Two scored examples. Verdicts follow the family-specific matching rules; \ding{51} = target recognized ($r_i{=}1$), \ding{55} = not recognized ($r_i{=}0$).}
\label{tab:app-cases}
\small
\renewcommand{\arraystretch}{1.15}
\begin{tabularx}{\textwidth}{@{}>{\raggedright\arraybackslash}p{0.17\textwidth}>{\raggedright\arraybackslash}X@{}}
\toprule
\multicolumn{2}{@{}l}{\textbf{Example 1: E-ACT 0030 (Emergence, ASCII Code Transformation)}} \\
\midrule
Image & \begin{minipage}[t]{0.40\textwidth}\vspace{0pt}\includegraphics[width=\linewidth]{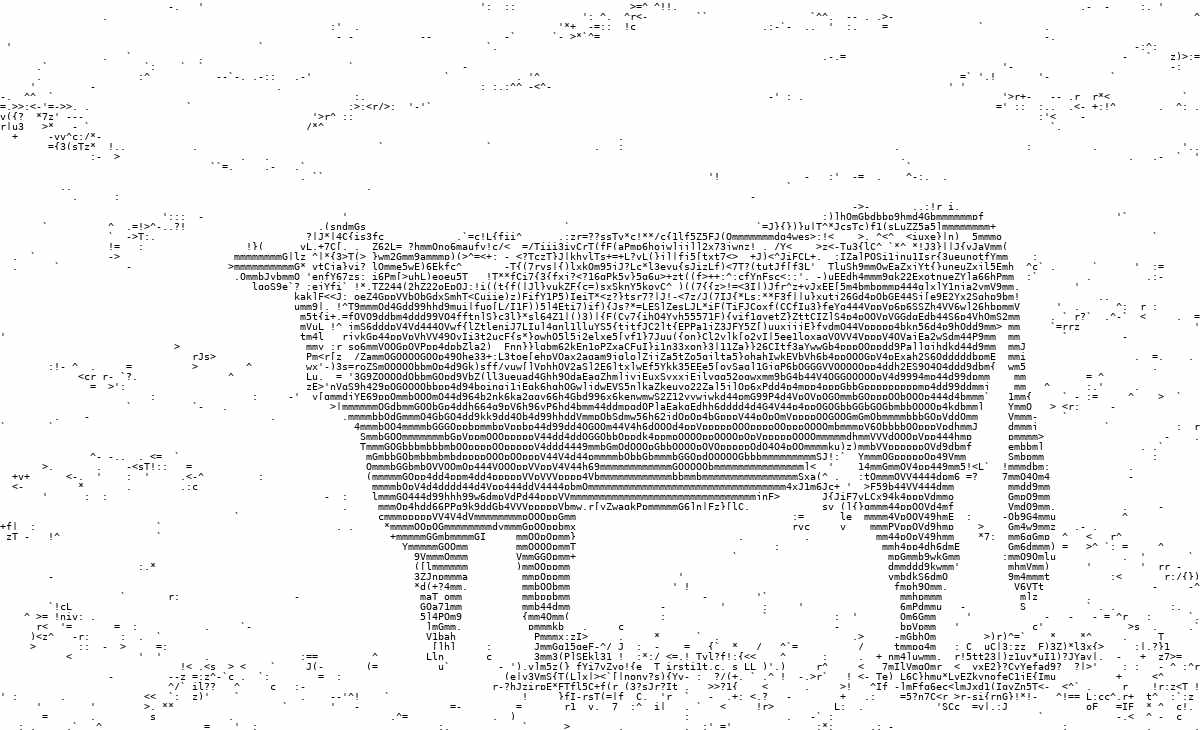}\end{minipage}\hfill
\begin{minipage}[t]{0.39\textwidth}\vspace{0pt}\raggedright \emph{Construction.} Inherited transformation (R1) of the OA photograph OA-0107: the photograph is contrast-stretched, gamma-adjusted, sharpened, and rendered with a character atlas, so the animal is visible only through the spatial density of ASCII-like symbols.\end{minipage} \\
Prompt & ``The picture is drawn with text characters (ASCII art); ignore the individual characters and look at the overall shape formed by their density, as if viewed from a distance. What animal is in this image? Describe the image in detail.'' \\
Target / aliases & $t_i$ = \texttt{cow}; $\mathcal{A}_i$ = \{\texttt{cow}, \texttt{cows}, and the Chinese name of the animal (\emph{ni\'u})\}. \\
Matching rule & The target or an alias must appear as a complete word in the normalized response (Chinese aliases as substrings). \\
GPT-4o \ding{51} & ``The image is an ASCII art representation of a \textbf{cow}. The cow is formed using a variety of characters, including letters, numbers, and symbols. The body of the cow is densely packed with characters, creating a solid silhouette. The legs, head, and tail are distinct, with variations in character density providing detail. [\dots]'' \\
Why correct & The response names the target \texttt{cow} as a complete word and describes the animal's global silhouette (legs, head, tail), i.e., the symbols are integrated into the intended whole. \\
GLM-5V-Turbo \ding{55} & ``This image is \textbf{ASCII art} depicting a \textbf{human portrait} rendered entirely in black text characters against a white background. [\dots] \textbf{Subject}: A person with shoulder-length or longer hair, facing forward or slightly angled [\dots] The specific identity of the person depicted is ambiguous due to the abstract nature of ASCII rendering [\dots]'' \\
Why incorrect & The model correctly identifies the \emph{carrier} (ASCII art, character density) but organizes the dense region into the wrong global form (a human portrait). Neither \texttt{cow} nor any alias appears, so $r_i{=}0$. The same model reaches 88.24\% on E-BWSE but 0.00\% on E-ACT (Table~\ref{tab:main-results}). \\
\midrule
\multicolumn{2}{@{}l}{\textbf{Example 2: M-THW 0105 (Multistability, Text Hiding Words)}} \\
\midrule
Image & \begin{minipage}[t]{0.22\textwidth}\vspace{0pt}\includegraphics[width=\linewidth]{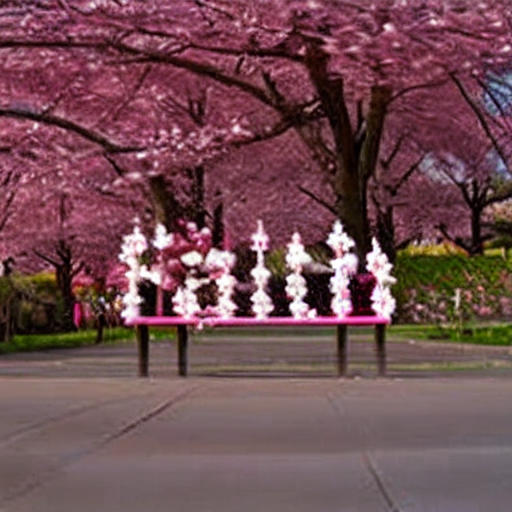}\end{minipage}\hfill
\begin{minipage}[t]{0.57\textwidth}\vspace{0pt}\raggedright \emph{Construction.} Template-conditioned diffusion (R3): the word \texttt{Faith} is rendered as a binary text mask and used as the ControlNet condition; the carrier scene (a park bench under cherry blossoms) is generated around it, so the letters are formed by the white ornaments on the bench.\end{minipage} \\
Prompt & ``The scene's elements are arranged so that, at a large scale, they also form letters; view the whole image as if from a distance or while squinting. What English word is in this image? Describe the image in detail.'' \\
Target / aliases & $t_i$ = \texttt{Faith}; $\mathcal{A}_i=\emptyset$ (the 20 M-THW words have no aliases). \\
Matching rule & The response must report explicit candidate text (e.g., quoted, or introduced by \emph{reads}/\emph{spells}) that equals the target after case and punctuation normalization; describing the scene alone gives no credit. \\
GPT-4o \ding{51} & ``The image depicts a serene outdoor scene featuring cherry blossom trees in full bloom. [\dots] In the foreground, there is a pink bench adorned with decorations that \textbf{spell out the word ``Faith''} in white letters. [\dots]'' \\
Why correct & The hidden word is reported as explicit text (``spell out the word `Faith'\,'') and equals $t_i$; the model separates the embedded word from the competing carrier interpretation. \\
GLM-5V-Turbo \ding{55} & ``This image depicts a \textbf{springtime scene} featuring: [\dots] A \textbf{bright pink bench} positioned in the middle of the frame. On the bench sit \textbf{seven white, ornate objects} resembling decorative chess pieces or elaborate figurines with intricate, stacked designs [\dots]'' \\
Why incorrect & The response gives an accurate carrier-only description: the letter strokes are perceived as ``ornate objects'' rather than grouped into a word, and no candidate text is reported, so $r_i{=}0$. This is the carrier-dominant failure illustrated in Figure~\ref{fig:overview}. \\
\bottomrule
\end{tabularx}
\end{table*}

\section{Data Construction Details}
\label{app:construction}

This section gives the per-family construction details. Every item is annotated with its family, category, language (Chinese or English), frozen target $t_i$, and alias set $\mathcal{A}_i$; additional family-specific annotations are described below.

\textbf{Reification.}
\emph{R-FSPO} (105 items, all Chinese, about $1300{\times}700$\,px): publicly available text images are processed by a script that crops or masks the secondary strokes or components of the characters; the target is the complete phrase (102 distinct phrases).
\emph{R-CCE} (99 of 117 candidates) and \emph{R-CCH} (39 of 40) are collected online composite-character images in which a Chinese four-character idiom is fused into a single glyph; each idiom is also annotated with its pinyin. R-CCE contains conventional glyphs built by simple cropping, splitting, or local composition; R-CCH contains artistic glyphs with heavier fusion and deformation (all $528{\times}528$\,px).

\textbf{Emergence.}
\emph{E-ACT} (321 items, 18 animal categories) and \emph{E-BWSE} (306 items, 19 categories) re-encode OA photographs, and each item is linked to its source OA photograph. Symbol encoding applies adaptive contrast stretching, gamma adjustment, and mild sharpening, and renders the result with a character atlas (output width 1{,}200\,px). Black--white encoding applies grayscale conversion, contrast enhancement, thresholding, and structured sparse deletion.
\emph{E-CBS} (117 of 118 candidates) collects Ishihara-style plates whose targets are digits (63), objects (45), words (7), or letters (2); for object targets, the annotation lists the groups of English keywords that must all be named.
\emph{E-DSV} (200 items: 100 Chinese idioms, 100 English idiomatic phrases, mostly $1024{\times}1024$\,px) composes a foreground text segment with a cast-shadow segment, which together form the target (e.g., \texttt{break} in the foreground and \texttt{the ice} in the shadow for \texttt{break the ice}). 

\textbf{Invariance.}
\emph{I-CCM} (214 items: 107 Chinese phrases, 107 English sentences) horizontally mirrors source text images; the target is the original (unmirrored) text, e.g., \texttt{Morning light crossed the empty platform.} 
\emph{I-AP} (328 items, 249 distinct identities) collects stylized, abstract, or deformed portraits of public figures from online sources; aliases contain documented name variants, transliterations, and Chinese names, and portraits of the same person are grouped under one identity.

\textbf{Multistability.}
\emph{M-THW} uses 20 English words (\texttt{Bold, Brave, Calm, Cure, Dream, Faith, Free, Grace, Grow, Hope, Joy, Kind, Light, Love, Peace, Smile, Trust, Truth, Unity, Wise}) with 20 candidates per word (400 candidates, $512{\times}512$\,px); six all-black candidates are removed, leaving 394.
\emph{M-THP} uses 114 short English phrases (e.g., \texttt{Be careful}, \texttt{Hold on}), one image each.
\emph{M-HAI} uses 12 animal templates (two each of bird, cat, dog, mouse, pig, and sheep), 16 carrier scenes (autumn leaves, canyon, desert, dining room, forest path (painting), gym, harbor (oil painting), icebergs, laboratory, leaves, music room (painting), rock cave, ruins, snow mountain, stones, train station (painting)), and three retained blending levels (levels 1, 3, and 5, with 192 images each; a higher level integrates the animal more fully into the scene), giving $12{\times}16{\times}3=576$ images ($512{\times}512$\,px).

\textbf{Generation model and prompts (R3).}
All three Multistability families are produced with ControlNet~\cite{zhang2023addingconditionalcontroltexttoimage} following Illusion Diffusion~\cite{burgert2023diffusionillusionshidingimages}: text targets are rendered as binary masks and animal templates are used directly as the control image, while a text prompt describes the carrier scene. For M-HAI the carrier prompt is built from the scene name (e.g., ``harbor, oil painting''). The 960 M-HAI candidates in Table~\ref{tab:construction-retention} correspond to $12{\times}16{\times}5$ blending levels, of which levels 1, 3, and 5 are retained.

\textbf{Original Animals (OA).}
OA keeps the 800 highest-scoring of 1{,}740 photographs from the KTH animal-image collection across 19 categories (panda 64, tiger 55, bear 54, cow 50, giraffe 50, goat 48, gorilla 46, cougar 45, horse 43, leopard 42, coyote 41, elephant 40, deer 39, zebra 37, sheep 32, skunk 32, penguin 29, kangaroo 28, lion 25). The quality score combines contrast, edge clarity, sharpness, subject ratio, color diversity, and saturation. 

\textbf{Screening and annotation.}
Annotators screen every candidate for target recognizability and visual quality (Section~\ref{sec:construction}). Targets and aliases are normalized with Unicode NFKC, case folding, and whitespace/punctuation normalization, and are frozen before any model is queried; no alias is added from model outputs.

\textbf{Sources and licenses.}
The benchmark code (scoring, querying, and figure scripts) is released under the Apache License 2.0. The images come from three kinds of sources: (i) images generated or transformed by us (E-ACT, E-BWSE, R-FSPO, E-DSV, I-CCM, M-THW, M-THP, M-HAI); (ii) OA photographs from the public KTH animal-image collection; and (iii) publicly available online images that we collected and curated (R-CCE, R-CCH, E-CBS, I-AP) and the source text images of the scripted families. Rights in third-party images remain with their owners, and the code license does not grant rights to these images; the images are distributed for non-commercial research on perceptual organization and model evaluation only, with metadata removed. I-AP contains stylized portraits of public figures and must not be used to identify, profile, or make inferences about individuals. 

\section{Experimental Details}
\label{app:exp-details}

\textbf{Querying and decoding.}
API-based VLMs are queried through OpenAI-compatible chat-completion endpoints with a single-turn user message containing the family-specific prompt and the image, sent as the original file in a base64 data URL; no system prompt, demonstration, or tool is used. Each prompt consists of a construction hint, an answer-type question, and the closing instruction \emph{``Describe the image in detail.''}, in this order. Table~\ref{tab:app-prompts} lists the hints and questions; the hint describes how the image is built and the question names the expected answer type, and neither reveals the target. Decoding is shared by all models: the temperature is 0.2, the output is limited to 1024 new tokens, and top-$p$ and top-$k$ are left at each endpoint's default. Each item receives one response, without voting or resampling. Requests time out after 180s; in the released querying script, transport-level failures (network or API errors) are retried up to three times with exponential back-off, content is never resampled, and an item whose request still fails is scored as a failure.

\begin{table}[h]

\centering
\caption{Family-specific prompts. Each prompt is the construction hint, followed by the answer-type question and the shared closing instruction \emph{``Describe the image in detail.''}}
\label{tab:app-prompts}
\small
\renewcommand{\arraystretch}{1.15}
\begin{tabularx}{\textwidth}{@{}l>{\raggedright\arraybackslash}X>{\raggedright\arraybackslash}X@{}}
\toprule
Family & Construction hint & Answer-type question \\
\midrule
R-FSPO & Parts of the characters' strokes are covered or removed; mentally complete the missing strokes from the remaining structure. & What Chinese text is written in this image? Give the complete text. \\
R-CCE, R-CCH & Several Chinese characters are fused into one combined glyph; separate its components and read them as a complete idiom. & What Chinese idiom is written in this image? Give the complete idiom in Chinese characters. \\
E-ACT & The picture is drawn with text characters (ASCII art); ignore the individual characters and look at the overall shape formed by their density, as if viewed from a distance. & What animal is in this image? \\
E-BWSE & The picture is a high-contrast black-and-white rendering; group the black and white patches into one overall shape. & What animal is in this image? \\
E-CBS & This is a colour-vision test plate; group the dots of similar colour to find the figure they form. & What number, letter, word, or object is shown in this image? \\
E-DSV (Chinese) & The visible text and its cast shadow show different parts of the target; combine the foreground text and the shadow text into one reading. & What Chinese idiom does this image show? Give the complete idiom in Chinese characters. \\
E-DSV (English) & Same as E-DSV (Chinese). & What English phrase does this image show? Give the complete phrase. \\
I-CCM (Chinese) & The text is mirrored (horizontally flipped); mentally flip it back before reading. & What Chinese text is written in this image? Give the complete text in Chinese characters. \\
I-CCM (English) & Same as I-CCM (Chinese). & What English sentence is written in this image? Give the complete sentence. \\
I-AP & This is a stylized, abstract or caricatured portrait of a real, well-known person; focus on the identity cues rather than the drawing style. & Who is the person in this image? Give the most specific name you can. \\
M-THW & The scene's elements are arranged so that, at a large scale, they also form letters; view the whole image as if from a distance or while squinting. & What English word is in this image? \\
M-THP & Same as M-THW. & What English phrase is in this image? \\
M-HAI & The scene's elements are arranged so that, at a large scale, they also form the shape of an animal; view the whole image as if from a distance or while squinting. & What animal is in this image? \\
\midrule
OA (reference) & Look at the main subject of the photograph. & What animal is in this image? \\
\bottomrule
\end{tabularx}

\end{table}

\textbf{Environment.}
All models are accessed through their providers' APIs, so the evaluation requires no local GPU. Querying and scoring are implemented in Python ($\geq$3.9) without third-party dependencies. The analysis and figures were produced on a CPU server (AMD EPYC 9654, 64 logical cores, 130\,GB RAM, Linux 6.8) with Python 3.14.6, NumPy 2.4.6, pandas 3.0.3, Matplotlib 3.11.0, and pyCirclize 1.10.1.

\textbf{Evaluation scale.}
Each model is evaluated on all 3{,}613 items (2{,}813 PO items and 800 OA items), one request per item; the 11-model evaluation therefore comprises 39{,}743 responses.

\end{document}